\documentclass[preprint,12pt]{elsarticle}

\usepackage{amssymb}
\usepackage{tikz}
\usetikzlibrary{automata}
\usepackage{pgfplots}
\journal{Neurocomputing}

\begin{document}

\begin{frontmatter}

%% Title, authors and addresses

%% use the tnoteref command within \title for footnotes;
%% use the tnotetext command for theassociated footnote;
%% use the fnref command within \author or \address for footnotes;
%% use the fntext command for theassociated footnote;
%% use the corref command within \author for corresponding author footnotes;
%% use the cortext command for theassociated footnote;
%% use the ead command for the email address,
%% and the form \ead[url] for the home page:
%% \title{Title\tnoteref{label1}}
%% \tnotetext[label1]{}
%% \author{Name\corref{cor1}\fnref{label2}}
%% \ead{email address}
%% \ead[url]{home page}
%% \fntext[label2]{}
%% \cortext[cor1]{}
%% \affiliation{organization={},
%%             addressline={},
%%             city={},
%%             postcode={},
%%             state={},
%%             country={}}
%% \fntext[label3]{}

\title{High-resolution semantically-consistent image-to-image translation}

%% use optional labels to link authors explicitly to addresses:
%% \author[label1,label2]{}
%% \affiliation[label1]{organization={},
%%             addressline={},
%%             city={},
%%             postcode={},
%%             state={},
%%             country={}}
%%
%% \affiliation[label2]{organization={},
%%             addressline={},
%%             city={},
%%             postcode={},
%%             state={},
%%             country={}}

\author[label1]{Mikhail Sokolov}
\ead{sokolov.usmk@gmail.com}
\affiliation[label1]{organization={Department of Applied Computer Science, University of Winnipeg},%Department and Organization
            addressline={515 Portage Ave}, 
            city={Winnipeg},
            postcode={R3B 2E9}, 
            state={Manitoba},
            country={Canada}}
            
\author[label1]{Christopher Henry}
\ead{ch.henry@uwinnipeg.ca}

\author[label2]{Joni Storie}
\ead{j.storie@uwinnipeg.ca}
\affiliation[label2]{organization={Department of Geography, University of Winnipeg},%Department and Organization
            addressline={515 Portage Ave}, 
            city={Winnipeg},
            postcode={R3B 2E9}, 
            state={Manitoba},
            country={Canada}}
            
\author[label2]{Christopher Storie}
\ead{c.storie@uwinnipeg.ca}

\author[label3]{Victor Alhassan}
\ead{victor.alhassan@nrcan-rncan.gc.ca}

\affiliation[label3]{organization={Canada Centre for Mapping and Earth Observation, Natural Resources Canada},%Department and Organization
            addressline={212 - 50 Place de la Cit$\acute{e}$, P.O. Box 162, 2nd Floor}, 
            city={Sherbrooke},
            postcode={J1H 4G9}, 
            state={Quebec},
            country={Canada}}

\author[label3]{Mathieu Turgeon-Pelchat}
\ead{mathieu.turgeon-pelchat@nrcan-rncan.gc.ca }

\begin{abstract}
Deep learning has become one of remote sensing scientists' most efficient computer vision tools in recent years. However, the lack of training labels for the remote sensing datasets means that scientists need to solve the domain adaptation problem to narrow the discrepancy between satellite image datasets. As a result, image segmentation models that are then trained, could better generalize and use an existing set of labels instead of requiring new ones. This work proposes an unsupervised domain adaptation model that preserves semantic consistency and per-pixel quality for the images during the style-transferring phase. This paper's major contribution is proposing the improved architecture of the SemI2I model, which significantly boosts the proposed model's performance and makes it competitive with the state-of-the-art CyCADA model. A second contribution is testing the CyCADA model on the remote sensing multi-band datasets such as WorldView-2 and SPOT-6. The proposed model preserves semantic consistency and per-pixel quality for the images during the style-transferring phase. Thus, the semantic segmentation model, trained on the adapted images, shows substantial performance gain compared to the SemI2I model and reaches similar results as the state-of-the-art CyCADA model. The future development of the proposed method could include ecological domain transfer, {\em a priori} evaluation of dataset quality in terms of data distribution, or exploration of the inner architecture of the domain adaptation model.

\end{abstract}

%%Graphical abstract
%\begin{graphicalabstract}
%\includegraphics[width=\textwidth,keepaspectratio]{images/generator.eps}
%\end{graphicalabstract}

%%Research highlights
%\begin{highlights}
%\item Research highlight 1: method to improve SemI2I model architecture to significantly boosts the model’s performance
%\item Research highlight 2: testing the CyCADA model on the remote sensing multi-band datasets
%\item Research highlight 3: achieving better accuracy of LULC maps generated using UDA models
%\end{highlights}

\begin{keyword}
unsupervised \sep domain \sep adaptation \sep remote \sep sensing \sep deep \sep learning \sep worldview-2 \sep spot-6 
%% keywords here, in the form: keyword \sep keyword

%% PACS codes here, in the form: \PACS code \sep code

%% MSC codes here, in the form: \MSC code \sep code
%% or \MSC[2008] code \sep code (2000 is the default)

\end{keyword}

\end{frontmatter}

%% \linenumbers

%% main text
\section{Introduction}
\label{section:intro}
\citet{rs_bigdata} states that in an era of big earth data, also called remote sensing (RS) big data, there are significant challenges associated with the high dependency on large-scale supervised land cover labels needed to generate effective map products. The use of an unsupervised domain adaptation method to align the representation of satellite images taken from different satellites and geo-graphical regions offers a solution to this challenge. The proposed approach reduces the discrepancy between the source domain (images with corresponding semantic labels) and the target domain (images without labels) so the segmentation model which is trained after can work effectively with both datasets. The key feature of this proposed method is that the adapted images are highly accurate in terms of semantic consistency, {\em i.e.} the objects in the adapted images preserve their original logical meaning. Semantically consistent adaptation is crucial in RS because each pixel brings certain information which should be preserved. This adaption method can be successfully incorporated in existing map production pipelines when the land cover labels for one area are missing, when the sensor resolution characteristics are different, or when both of these situations occur at once. The presented solution is increasingly important for government and private industries since training labels in RS is limited or not publicly available, and expensive to obtain.

{\em Land use and land cover} (LULC) maps are generated products resulting from satellite imagery that relate each pixel in the satellite image to a specific class of objects, {\em e.g.}, vegetation, hydro, road, {\em etc.} Land use applications generally serve to monitor the changes in human economic and cultural activity on the land ({\em i.e.}, recreation, agricultural, mining, {\em etc.}). Land cover, in turn, refers to which natural (rivers, forests, snow, {\em etc.}) or human-made objects (buildings, cars, roads, {\em etc.}) which exists on the ground \citep{nrcan}. It is land cover features that are detected using reflected energy recorded per pixel while the land use is inferred based on land cover elements.

Governments and commercial organizations involved in land management widely use LULC maps because they can provide valuable and accurate information if generated on a regular basis. For example, LULC maps are used for emergency response to efficiently deploy restoration forces after significant flooding or landslides. Demand for up-to-date LULC maps is increasing because there are more satellite platforms and sensors on those platforms, providing big RS data for more information and variation of its use \citep{rs_bigdata}. Furthermore, new constellations that provide data at higher temporal  frequencies ({\em i.e.} shorter revisit periods) and broader area coverage are coming online \citep{newsat}.

The process of generating LULC maps involves many trained specialists. Before deep learning was introduced for RS data, the process of LULC map production was semi-automated. It required a human in the loop, but also made use of existing tools and algorithms \citep{lu_lulc}. To label one satellite image, a knowledgeable person needed to visually assess each image pixel and assign a corresponding label. This time consuming and expensive labour, coupled with the need for increased temporal frequency of LULC maps, drove the need for a fully automated solution. This need and the rise of deep learning algorithms \citep{lecun_nn} led to the development of segmentation algorithms for LULC map generation. 

Semantic segmentation is a process of assigning each pixel a logical label, {\em e.g.}, vegetation, road, background, {\em etc.} Recent achievements in image classification problems using convolutional neural networks (CNNs) led to significant advances in per-pixel segmentation tasks as well \cite{chen_deeplab, shen_efficient, liu_dpn}. This, in turn, was used in a wide spectrum of computer vision applications \cite{geiger_kitti, tsai_harmon}. The success of CNN-based segmentation algorithms led to their use in automated development of LULC maps, with very good results \citep{victor_is, story_lulc, segm_allh}. However, the process of annotating a large number of images needed for training CNN-based semantic segmentation models is a significant challenge. For example, the release of each new satellite (and corresponding sensors) usually requires the creation of a new LULC labelled training dataset due to spatial and spectral differences in the new sensors. To reduce costs of developing new training datasets, the ability to adapt models trained with labelled data from one domain (called the {\em source} domain) to another domain (called the {\em target} domain) would be very useful due to the expense of labelling new datasets. For example, different RS scene image datasets may be taken from different types of sensors, in different weather conditions, in different geographical areas, and have different resolutions and scales. Consequently, the domain distribution discrepancy may be significant from one dataset to another, which makes models trained on a source dataset not useful for new target domains.

To tackle this issue, researchers are investigating and have been developing domain adaptation (DA) techniques that are used to close the gap between source and target domains. Most common approaches aim to align features across two domains so that a semantic segmentation model can generalize across them \citep{goodfellow_gan, tsai_adaptsegnet, liu_kl, liu_curves}. However, compared to classification tasks, feature adaptation in segmentation tasks is more complicated because the model has to encode a diversity of different visual characteristics such as appearance, shapes and context \citep{tsai_adaptsegnet}. Another group of adaptation methods deal with a style transferring task and has shown excellent performance when applied to RS datasets \citep{tasar_colormapgan, tasar_i2i, tasar_daugnet}.

Given the DA methods which utilize style-transferring approaches raises the question of which method is more efficient and thus is worth pursuing further development. The dominance of the CycleGAN-based methods \citep{zhu_cyclegan} is challenged by the models with the adaptive instance normalization layer embedded; for example, the SemI2I model proposed by \citet{tasar_i2i}. This model is focused on preserving the semantic consistency during the style-transferring part, which is crucial in RS. However, SemI2I architecture is not sufficiently sophisticated to provide high-quality image-to-image translation outputs because, initially, it was designed to reduced memory consumption and computation speed. 

This paper presents a new unsupervised domain adaptation method called {\em High-resolution semantically-consistent image-to-image translation} (HRSemI2I) that employs the AdaIN \citep{huang_adain} layer and aims to transfer the target domain's style to the source images preserving semantic consistency and per-pixel image quality. This research proposes to significantly improve the architecture of the SemI2I model to boost model performance and make it competitive with the state-of-the-art CyCADA model \citep{hoffman_cycada}. We also test the CyCADA model on the RS multi-band datasets which has not been applied to this application to date. CyCADA was initially trained on the synthetic photorealistic datasets, such as GTA5 and Synthia, and validated on the Cityscapes dataset of vehicle-egocentric real-world images.

The next section summarizes the research related to unsupervised domain adaptation methods, in general and for RS data. This is followed by a discussion of proposed network architecture and training methods including details of the source and target datasets as well as the experiments conducted.

\section{Related Works}
\label{section:related}
Domain adaptation techniques are used to close the gap between source and target domains. They became popular after \citet{ganin_da, ganin_da2} proposed unsupervised domain adaptation through backpropagation. Since then, many variants and model architectures that fit the domain adaptation task have been proposed. The most common approaches aim to align features across two domains by using a semantic segmentation model that can generalize the two domains. For instance, a feature-level adaptation method was proposed by \citet{tzeng_da}, where he uses so-called domain confusion loss, which directly minimizes the distance between source and target representations, thus initializing domain invariance. Also, there is a classification loss which solves the image classification task. 

\citet{hong_condgen} proposed using a conditional generative adversarial network for structured domain adaptation. The central part of the proposed model is a conditional generator which is aimed to enhance source domain features to have a similar distribution as the target. The conditional generator consists of several residual blocks. There is also a CNN encoder with five convolutional layers that extracts target features. The target domain features and enhanced source domain features are then passed through the discriminator represented by a multi-layer perceptron. Thus, the domain adaptation task is resolved via adversarial learning. At the same time, the semantic segmentation task is solved after the cross-entropy loss is calculated by passing the enhanced source features through the deconvolutional layer. 

Another feature-level adaptation method was proposed by \citet{tzeng_da2}, which combines adversarial learning with discriminative feature learning. In particular, during the training, the model must learn a discriminative mapping of target images to the source feature space. The model consists of a CNN feature encoder and an FCC discriminator, and the domain adaptation is solved via adversarial learning. The methods mentioned above are limited by a common issue - they do not enforce semantic consistency while aligning feature representation of both domains. However, this is crucial when the final goal is the semantic segmentation of target samples. The following pixel-level domain adaptation methods were developed to resolve this issue. 

Originally based on CycleGAN \cite{zhu_cyclegan} and feature-level adaptation methods proposed in \cite{ganin_da, tzeng_da2}, the CyCADA model was proposed by \citet{hoffman_cycada} to preserve semantic consistency through pixel-level unsupervised domain adaptation. The model uses a noisy labeller, which is a semantic segmentation model trained on source data and applied to target data without adaptation. After that, a pixel-space adaptation process is performed using two generators and two discriminators. The first generator and discriminator work to translate the target domain feature representation to source samples and the second pair of generator and discriminator work as a whole to translate the source domain feature representation to target samples. The noisy labeller, trained previously, is integrated into the training process and encourages an image to be classified in the same way after translation as it was before translation, according to this classifier. Finally, the semantic segmentation model with feature-level adaptation is trained on source samples stylized as target ones. 

Another pixel-level adaptation method is aimed to preserve semantic consistency, which is called bidirectional learning, was proposed by \citet{li_da}. The authors explore a model where two separate modules cooperate. There is an image-to-image translation model and segmentation adaptation model, both similar to the ones proposed in \cite{hoffman_cycada}. The learning process involves two directions: ``translation-to-segmentation" and ``segmentation-to-translation", and that is why the whole process is a closed-loop cycle. Moreover, a self-supervised learning (SSL) approach is incorporated into a segmentation adaptation module. This way, different from segmentation models trained only on source data, the segmentation adaptation model uses both source and target images for training. Iteratively predicting labels for the target domain, they are considered as the approximation of the ground truth labels. Thus, only those with high probability are used in training over and over again. After the segmentation model is trained on the source and target labels, it is used as the noisy labeller discussed above, thus making the training process looped. Even though these two methods are both focused on the pixel-level adaptation task, where semantic consistency is preserved, the source and target datasets they were trained on are represented by a first-view car driving samples. The following pixel-space adaptation methods were applied to the RS imagery and are claimed to be effective. 

In \citep{liu_curves}, Liu et al. proposed using a curve feature extractor to represent each pixel of the input image as a curve. A semantic segmentation model DeepLab v3+ \citep{chen_dl3plus}, pretrained on the source images, is used to extract deep features for both domains. After the deep features are extracted, each pixel in the input image is converted into a feature curve. Then, the conditional generative adversarial network (cGAN) aligns the representation of the curves so that they are indistinguishable for the discriminator model regarding the source and  target domains. This way, the discrepancy between the domains is reduced. 

Another method was proposed by Liu et al. in \citep{liu_kl} where the authors propose using Kullback–Leibler constraint in their domain adaptation framework (KL-ADDA). The model consists of a generator and discriminator parts. The generator is represented by the DeepLab v3+ framework, and the discriminator is a fully-convolutional neural network, similar to \citep{long_fcn}. First, the generator is pretrained on the source images with corresponding labels. Then, the images from both domains are passed through the generator, and the semantic labels are acquired, where a semantic loss for the source domain is calculated. After this, predicted labels are passed to the discriminator, which decides what domain they belong to; thus, the discriminator loss is calculated. Afterwards, a KL divergence loss is calculated using the intermediate features extracted by the discriminator. Finally, the adversarial loss is calculated based on the discriminator loss and the KL divergence loss and is forwarded as a constrain for the generator. Even though this method represents the improved \citep{tsai_adaptsegnet}, it still lacks semantic consistency preservation.

The whole group of the pixel-level adaptation methods which were validated on RS imagery were proposed by \citet{tasar_colormapgan, tasar_i2i, tasar_daugnet}, where the authors put special attention to the semantic consistency preservation. The first method is called ColorMapGAN and is aimed to linearly shift each source sample band distribution to match the target domain distribution. At first, a U-Net \citep{ronneberger_unet} classifier is trained only on source samples and corresponding labels. After that, a ColorMapGAN module is trained to transfer the visual appearance of the source samples to look like the target ones. The ColorMapGAN consists of a generator and a discriminator. The generator is an architecturally simple construction that is represented by only scaling and shifting matrices. Thus, each source sample band (red, green, and blue) is passed through the scaling and shifting operations. The output of such transformations is then passed through the discriminator, which is similar to \citep{long_fcn}. The discriminator decides how close the transformed source sample is to the target distribution. The generator's goal is to fool the discriminator by faking the source images. It is crucial to notice here that the semantic consistency is preserved during such transformations because there are neither convolutional nor pooling layers. The final step of the training process is to fine-tune the initial classifier with faked source samples and their corresponding labels. 

The drawback of the ColorMapGAN is that it processes each band separately, which generates slightly noisy outputs. Another model proposed by these authors is called DAugNet. There is only one image-to-image translation part for the source-to-target and target-to-source style directions. There is also only one discriminator which estimates how accurate the style translation was. It can be used to evaluate both direction translations because it has domain-specific output layers. The scaling and shifting of feature representation are performed by constant predefined target-style vectors for the mean and variance, which do not change during the training. The same losses enforce the semantic consistency of the transformed samples as in the previously discussed method. After the style transferring part is done, the target-like source samples are used for semantic segmentation model training and validation on the target dataset. 

The last method that was developed with application to RS data is called semantically consistent image-to-image translation (SemI2I). The idea of this method is close to \citep{zhu_cyclegan} but has some specific differences. First, the image-to-image translation module operates using the AdaIN layer between the encoder and decoder part. During training, this layer scales and shifts the input source sample feature representation to match the target domain's accumulated mean and variance. Also, the image-to-image part consists of relatively shallow convolutional models; thus, it can be trained quickly. The following losses enforce semantic consistency of the translated source images: cross-reconstruction loss, self-reconstruction loss, and image gradient loss computed for the original source image and its translated to target domain representation copy. Additionally, the authors re-size the low-level features extracted by the first convolutional encoder layer and concatenate them to each deconvolution layer in the corresponding decoder. After the image-to-image part is trained, a semantic segmentation model is trained on translated source images with corresponding labels and validated on the target dataset. Inspired by \citep{tasar_i2i}, this paper proposes another unsupervised domain adaptation model which utilizes an AdaIN layer and is highly semantically consistent while performing the style-transferring task. The best building blocks from \citep{tasar_i2i} and \citep{ronneberger_unet} are taken and combined, so the resulting model achieves the state-of-the-art performance in the adaptation of the RS imagery.

\section{Methodology}
\label{section:methodology}
\subsection{Architecture}
\label{section:architecture}
Generative adversarial networks (GANs) are commonly used in unsupervised domain adaptation to close the gap between the source and target domains. In the proposed model, the goal of the generator {\em G} is to generate outputs that are indistinguishable from the real data. The discriminator {\em decides} how close the transformed source sample is to the target distribution. The generator's goal is to fool the discriminator by producing fake target images using source images as input. Thus, the domain adaptation task is resolved via adversarial learning.
\subsubsection{Generator}
\begin{figure}[!ht]
	\centering
	\includegraphics[width=\textwidth,keepaspectratio]{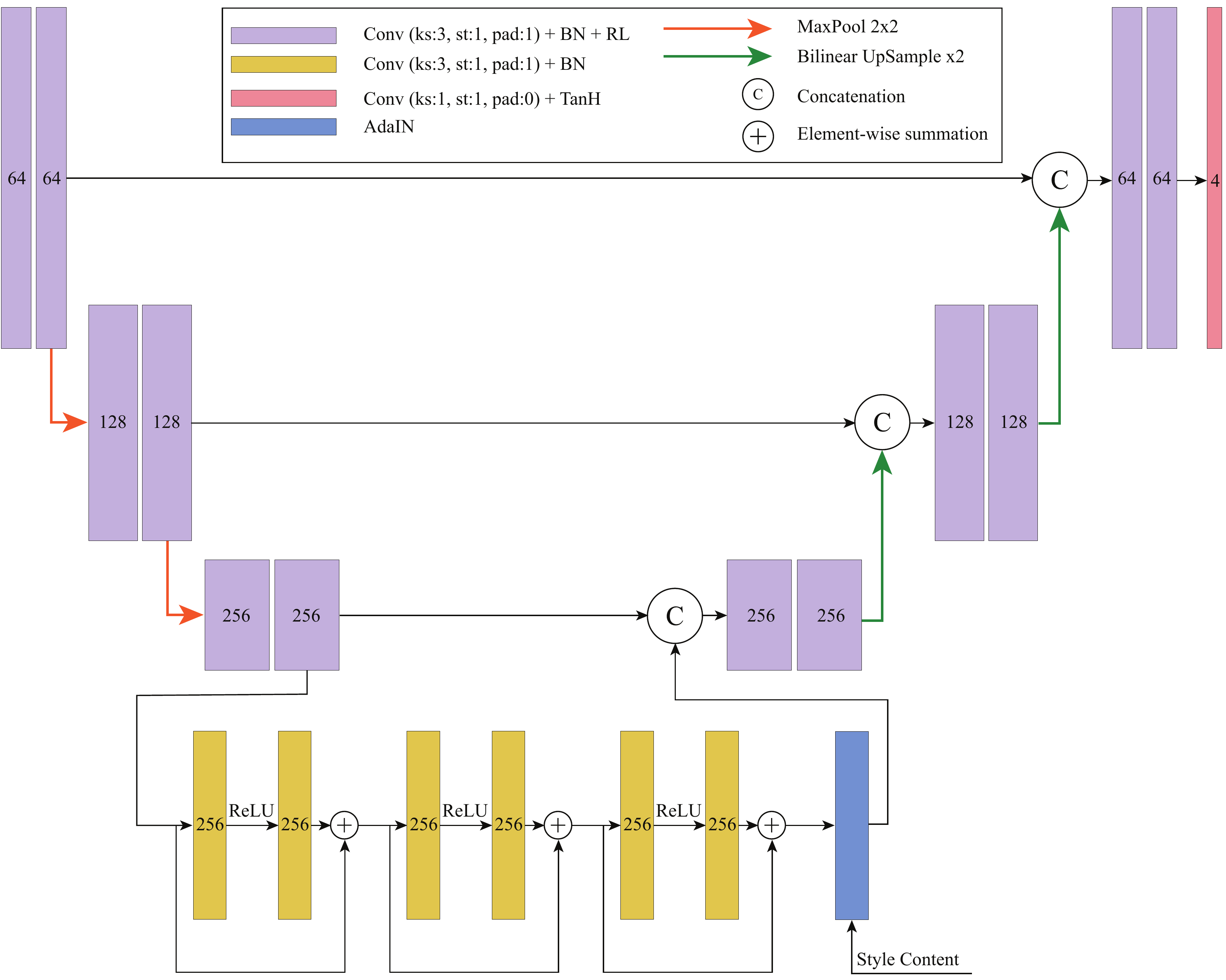}
	\caption{The architecture of the proposed generator, where $BN$, $RL$ and $TanH$ designate batch normalization layer, ReLU activation layer, and hyperbolic tangent activation, respectively.}
	\label{fig:figgen}
\end{figure}

In the proposed method, the generator, $G$, is represented by a feature encoder, including repeating residual parts, an AdaIN layer, and a feature decoder. The feature encoder is represented by the first three convolutional blocks of the U-Net model. The channel number for each encoder's convolutional block is 64, 128, and 256, while the first convolutional block has four input channels to operate on 4-band multi-spectral images. The intermediate part of the {\em G} is represented by three residual blocks, each of which has 256 channels in the convolutional layer. The AdaIN layer is attached to the last layer of the last residual block. Given content input from the source images and style input from target images, the AdaIN layer adjusts the mean and variance of the content input, thus, forcing it to be similar to the style input in terms of the distribution. AdaIN does not have trainable parameters, and it adaptively calculates affine transformation parameters from the intended domain:
\begin{equation}\label{eq:adain}
AdaIN(c,s)=\sigma(s)\left( \frac{c-\mu(c)}{\sigma(c)}\right)+\mu (s),
\end{equation}
where $c$ is a content input, $s$ is a style input, $\sigma$ returns the variance of the input, and $\mu$ returns the mean of the input. As can be seen from the equation, this layer scales content input by $\sigma(s)$ and shifts it by $\mu(s)$. The decoder part mirrors the encoder architecture and has 256, 128, and 64 channels in 3 convolutional blocks, followed by the 4-channel output convolutional block with the hyperbolic tangent activation function. Also, skip connections between the convolutional blocks of the encoder and decoder are maintained. The overall generator's architecture is depicted in Fig. \ref{fig:figgen}. 
\subsubsection{Discriminator}
\begin{figure}[!ht]
	\centering
	\includegraphics[width=\textwidth,keepaspectratio]{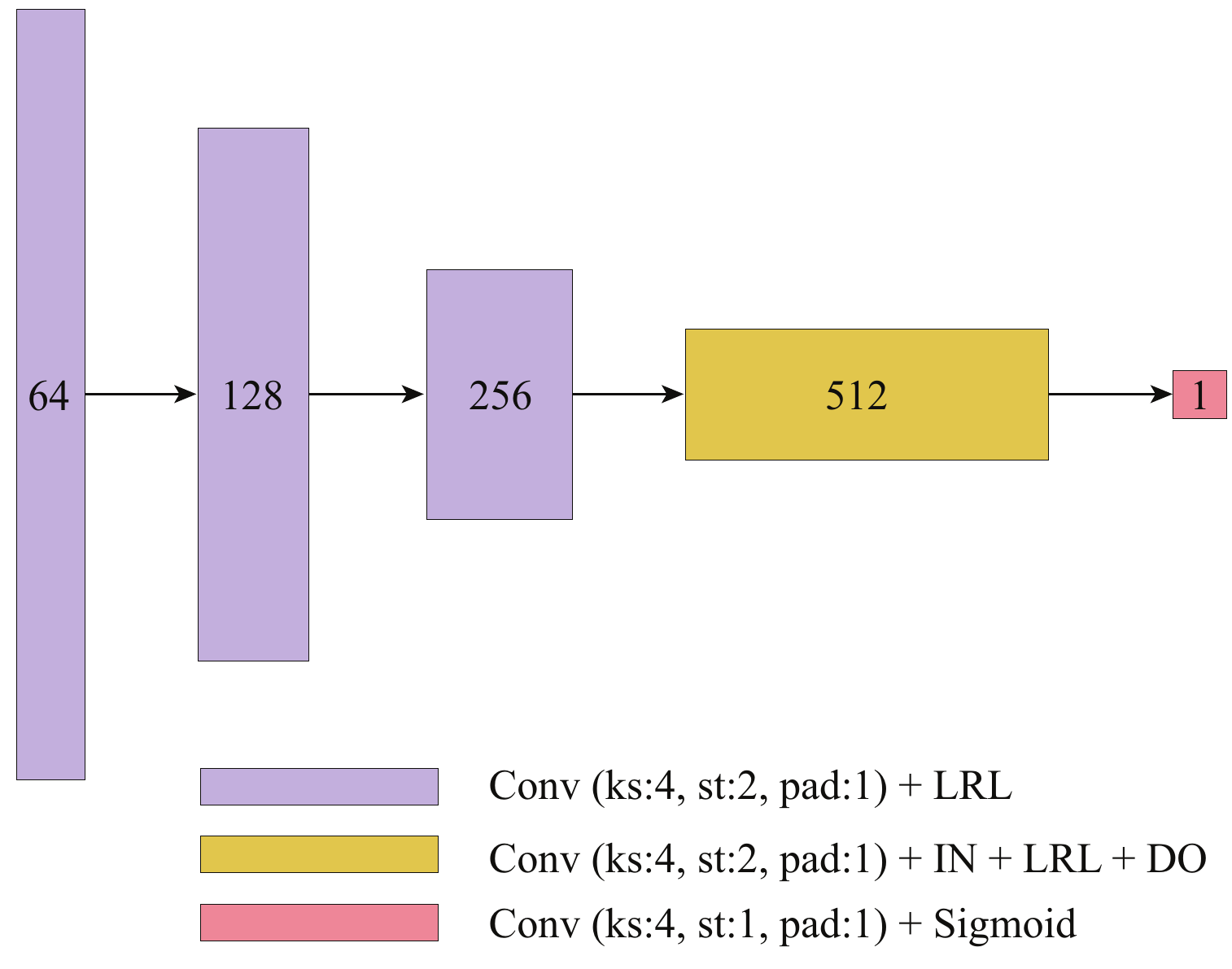}
	\caption{The architecture of the proposed discriminator, where $IN$, $LRL$, $DO$ and $Sigmoid$ designate instance normalization layer, leaky ReLU activation layer, dropout layer, and sigmoid activation, respectively.}
	\label{fig:figdisc}
\end{figure}

The discriminator part of the GAN consists of five convolutional layers, with the kernel size of 4$\times$4 and the stride parameter equal to 2. The number of channels for each layer is 64, 128, 256, and 512, followed by the output convolutional layer with the number of channels equal to 1 and the sigmoid activation function. The first four convolutional layers are followed by a leaky rectified linear unit with a coefficient of 0.2 and an instance normalization layer. The overall discriminator's architecture is depicted in Fig. \ref{fig:figdisc}. 
Altogether, the generator and discriminator are being optimized to minimize the following loss function:
\begin{equation}\label{eq:gan_loss}
L_{GAN}(D,G)=\mathbb{E}[\log D(x_t)]+\mathbb{E}[\log (1-D(G(x_s)))],
\end{equation}
where $x_t$ and $x_s$ are samples from the source and target domains, respectively. There are two pairs of a generator and a discriminator, and each pair is for the direction of style transformation: source-to-target and target-to-source.
%\subsubsection{Noisy Labeller}
%\label{section:noisy}
%A noisy labeller is used to improve the semantic consistency, similar to [?]. The noisy labeller is represented by the DeepLab v2 framework [?] with a modified input number of channels equal to 4. The labeller's weights stay frozen during the training process and do not update. This module works as a constraint to the generator preserving source content during the image-to-image translation process. The constrain can be expressed as the following loss function:
%\begin{equation}\label{eq:labeller}
%\begin{array}{l}
%L_{sem}(G_{S \rightarrow T}, G_{T \rightarrow S}, f_S)=L_{segm}(f_S, G_{S \rightarrow T}(X_S), p(f_S, X_S))\\
%+ L_{segm}(f_S, G_{T \rightarrow S}(X_T), p(f_S, X_T))
%\end{array}
%\end{equation}
%where $L_{segm}$ is a cross-entropy loss for the segmentation task, $f_S$ is the noisy labeller pretrained on the source images with corresponding labels, and $p(f_S, X)$ is a predicted label for the given input $X$ (either source or target).
\subsection{The method}
\label{section:method}
\subsubsection{Style transformation}
The style transformation part of the proposed method is achieved through the simultaneous work of the generator and discriminator assigned to each domain. The core of the transformation is hidden in the AdaIN layer and the adversarial training. Let $A$ (resp. $B$) represent a source (target) domain sample, and let $fake_A$ ($fake_B$) represent a target (source) sample transformed to the style of the source (target) domain. Then, to generate $fake_B$ and $fake_A$, the source image is passed through the encoder and residual blocks of the $G_{S \rightarrow T}$ and the target image is passed through the encoder and residual blocks of the $G_{T \rightarrow S}$. After the deep features for both domain inputs are extracted, the mean $\mu$ and variance $\sigma$ are calculated for each domain input. Since the mean and variance can significantly vary from input to input even within the same domain, the global mean and variance are used. This approach finds balanced values for each parameter through the accumulation process using the formulas:
\begin{equation}\label{eq:musigma}
\begin{array}{l}
\mu_{glob}=decay\_rate*\mu_{glob}+(1-decay\_rate)\times\mu_{c},\\
\sigma_{glob}=decay\_rate*\sigma_{glob}+(1-decay\_rate)\times\sigma_{c},
\end{array}
\end{equation}
where $\mu_{glob}$ and $\sigma_{glob}$ are the global mean and variance, and $\mu_{c}$ and $\sigma_{c}$ are mean and variance for the current input image batch. After current the $\mu_{c}$, $\sigma_{c}$ are calculated for both domains, the extracted features from each domain encoder are scaled and shifted and the resulting features are passed through the opposite domain decoder. Using the formulas, $fake_B$ and $fake_A$ can be expressed as
%
%%The style transformation part of the proposed method is achieved through the simultaneous work of the generator and discriminator assigned to each domain. The core of the transformation is hidden in the AdaIN layer and the adversarial training. Let $A$ and $B$ represent source and target domain samples, respectively. Then, to generate $fake_B$ and $fake_A$ (a source and target samples in the style of the target and source domains, respectively), the source image is passed through the encoder, and residual blocks of the $G_{S \rightarrow T}$ and the target image is passed through the encoder and residual blocks of the $G_{T \rightarrow S}$. After the deep features for both domain inputs are extracted, the mean $\mu$ and variance $\sigma$ are calculated for each domain input. Since the mean and variance can significantly vary from input to input even within the same domain, the global mean and variance are used. It allows to find the balanced values for each parameter through the accumulation process using the following formula:
\begin{equation}\label{eq:fakeAB}
\begin{array}{l}
fake_A=Dec_A(St(Enc_B(I_t), Enc_A(I_s))),\\
fake_B=Dec_B(St(Enc_A(I_s), Enc_B(I_t))),
\end{array}
\end{equation}
where $I_s$ and $I_t$ are the source and target samples, respectively, and $St()$ denotes the AdaIN layer. After that, assigned discriminators are used to evaluate how close the $fake_A$ and $fake_B$ are to the source and target domain distribution, respectively, the total adversarial loss for both generators is calculated, and the weights are optimized. Then, the discriminators' weights are also updated. 

\subsubsection{Semantic consistency}
The proposed method employs several constraints to enhance the semantic consistency of the style-transferring operation. It is a cross-reconstruction loss, which is an L1-norm of the original image and its reconstructed version. After the fake output is acquired, it is passed through the opposite generator to get a fake version of the faked original image. Ideally, the reconstructed image and the original one must be the same. The cross-reconstruction loss can be expressed as:
\begin{equation}\label{eq:crossrec}
\begin{array}{l}
L_{cross}=|I_{s}-Dec_A(St(Enc_B(fake_B), Enc_A(fake_A)))|\\
+|I_{t}-Dec_B(St(Enc_A(fake_A), Enc_B(fake_B)))|.
\end{array}
\end{equation}

Another constraint is known as self-reconstruction loss. After the embedding is extracted by the related encoder (before the AdaIN layer), it is passed through the same-domain decoder, and the L1-norm is calculated for the original image and its self-reconstructed version. The loss function can be expressed as:
\begin{equation}\label{eq:selfrec}
L_{self}=|I_{s}-Dec_A(Enc_A(I_s))| + |I_{t}-Dec_B(Enc_B(I_t))|.
\end{equation}

Also, a gradient loss is used as an additional semantic consistency constraint. After the fake version of the original image is generated, the first-order image derivative is calculated for both using a Sobel operator \citep{gauss}. Having the difference between them as small as possible, the model is forced to preserve the edges of the objects in the training images. The gradient loss can be expressed as:
\begin{equation}\label{eq:gradientloss}
L_{grad}=|grad(I_s) - grad(fake_B)| + |grad(I_t) - grad(fake_A)|,
\end{equation}
where $grad(\cdot)$ is a spatial gradient operator. 
%The last constraint which maintains the semantic consistency is a noisy labeller. As was shown in Sec. \ref{section:noisy}, it enforces the labels of the original images, and their style-transferred versions are as close as possible. The distance between them is calculated using a cross-entropy loss. The Eq. \ref{eq:labeller} can be re-written as:
%\begin{equation}\label{eq:noisysimle}
%L_{sem}= -\sum_{i=1}^{n} p(f_S, I_s)\cdot\log p(f_S, fake_B) - \sum_{i=1}^{n} p(f_S, I_t)\cdot\log p(f_S, fake_A),
%\end{equation}
%where $n$ is a number of classes in the segmentation task.

\subsubsection{Training}
Given the source domain images designated as $I_s$, and the target domain images denoted as $I_t$, the training process for the domain adaptation network can be expressed in the following iterations:
\begin{enumerate}
%\item The noisy labeller is trained on the $I_s$ and $Y_s$, then its weighs are frozen and are not updated during the training process.
\item The global mean and variance variables are initiated with zeros.
\item The inputs $I_s$ and $I_t$ are passed through the corresponding modules of $G_A$ and $G_B$, where $G_A$ and $G_B$ denote the generators assigned to the source-to-target and target-to-source transformations, respectively. 
\item Freeze the weights of the discriminators. Then, $fake_A$ and $fake_B$ are passed through the corresponding discriminator network, and the probabilities are acquired.
\item After the outputs are generated, the subsequent losses of the generators are calculated: the cross-reconstruction loss, the self-reconstruction loss, the gradient loss, and the adversarial loss of the generators.
\item The weights of the generators are updated.
\item Unfreeze (or activate) the discriminator weights. Then, $fake_A$ and $fake_B$ are passed through the corresponding discriminator networks, the probabilities are acquired, and the adversarial loss of the discriminators is calculated.
\item The weights of the discriminators are updated.
\item The global mean and variance are updated.
\item After the training process is done, all generators are saved along with the global mean and variance vectors.
\end{enumerate}

\subsubsection{Testing}
The testing stage uses only certain modules from each of the generators. To get the $fake_B$ output, which is a source domain image, transferred to the target domain style, the encoder part of the $G_A$ and the decoder part of the $G_B$ are needed together with the resulting global mean and variance parameters. The inference of the trained DA model can be expressed as:
\begin{equation}\label{eq:inference}
fake_B=Dec_B(St(Enc_A(I_s), (\mu_{glob}^{target}, \sigma_{glob}^{target}))),
\end{equation}
where $\mu_{glob}^{target}$, $\sigma_{glob}^{target}$ are the global mean and variance, calculated for the target domain inputs during the training stage.

\section{Experiments}
\label{section:experiments}
\subsection{The datasets}
\label{section:datasets}
The source and target datasets used in this work were acquired by WoldView-2 and SPOT-6 satellites, respectively. Both were automatically annotated by five semantic classes: background, vegetation, hydrology, roads, and buildings. The original satellite images were provided by Natural Resources Canada \citep{nrcan} as preprocessed rasters (GeoTIFFs) with corresponding labels in vector format (GeoPackage). 

\begin{figure}[!ht] 
\centering
\begin{tikzpicture} 
    \begin{axis}[
        ybar,
        xlabel = \empty,
        xmin = 0.5,
        xmax = 5.5,
        ymin = 0,
        ymax = 55,
        axis x line* = bottom,
        axis y line* = left,
        ylabel= \% pixels,
        width= 0.9\textwidth,
        height = 0.6\textwidth,
        ymajorgrids = true,
        bar width = 10mm,
        xticklabels = \empty,
        nodes near coords,
        extra x ticks = {1,2,3,4,5},
        extra x tick labels = {background, vegetarion, hydro, roads, buildings},
        ]
        \addplot+[mark=none, very thick] coordinates {
            (1,48.7)
            (2,38.2)
            (3,9.3)
            (4,1.8)
            (5,2.0)
        };
    \end{axis} 
\end{tikzpicture}
\caption[Label distribution in the source dataset.]{Label distribution in the source dataset.}
\label{fig:sourcelbl}
\end{figure}
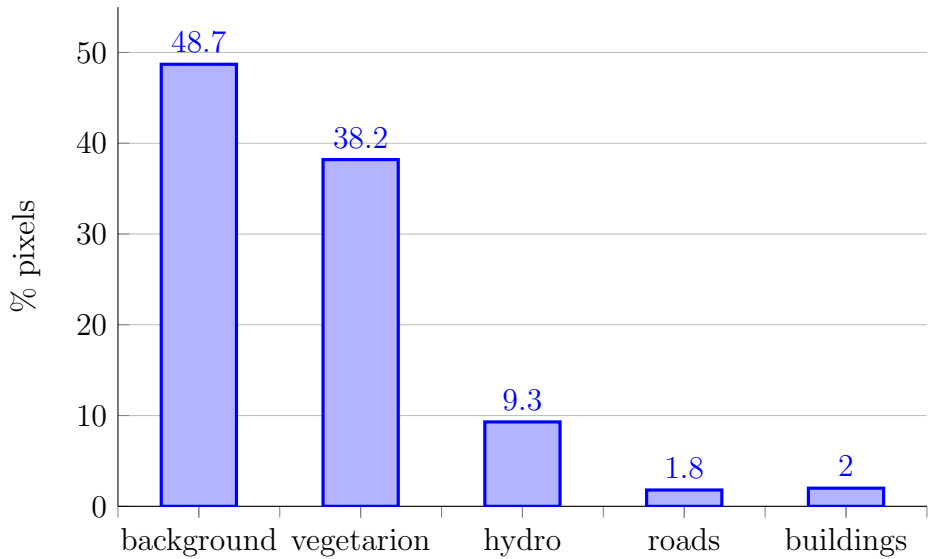

The domain with known labels (source) for this project is represented by the WoldView-2 imagery dataset, which is a preprocessed 0.5-meter spatial resolution multi-band 8-bit (resampled) set of images taken in spring, summer and fall months across Canada. The images were further downsampled to 1.5-meter spatial resolution to match the resolution of the target images. The GeoTIFF raster images and corresponding label geopackages were cropped into samples of size 4 $\times$ 512 $\times$ 512 and 1 $\times$ 512 $\times$ 512, respectively, where 4 represents the number of bands (blue, green, red, and near-infrared) in the image file and 1 represent a single band of the corresponding label file which was rasterized and saved in GeoTIFF format. The total number of source samples is 5560, with the label distribution represented in Fig. \ref{fig:sourcelbl}.

\begin{figure}[!ht]
	\centering
	\includegraphics[width=\textwidth,keepaspectratio]{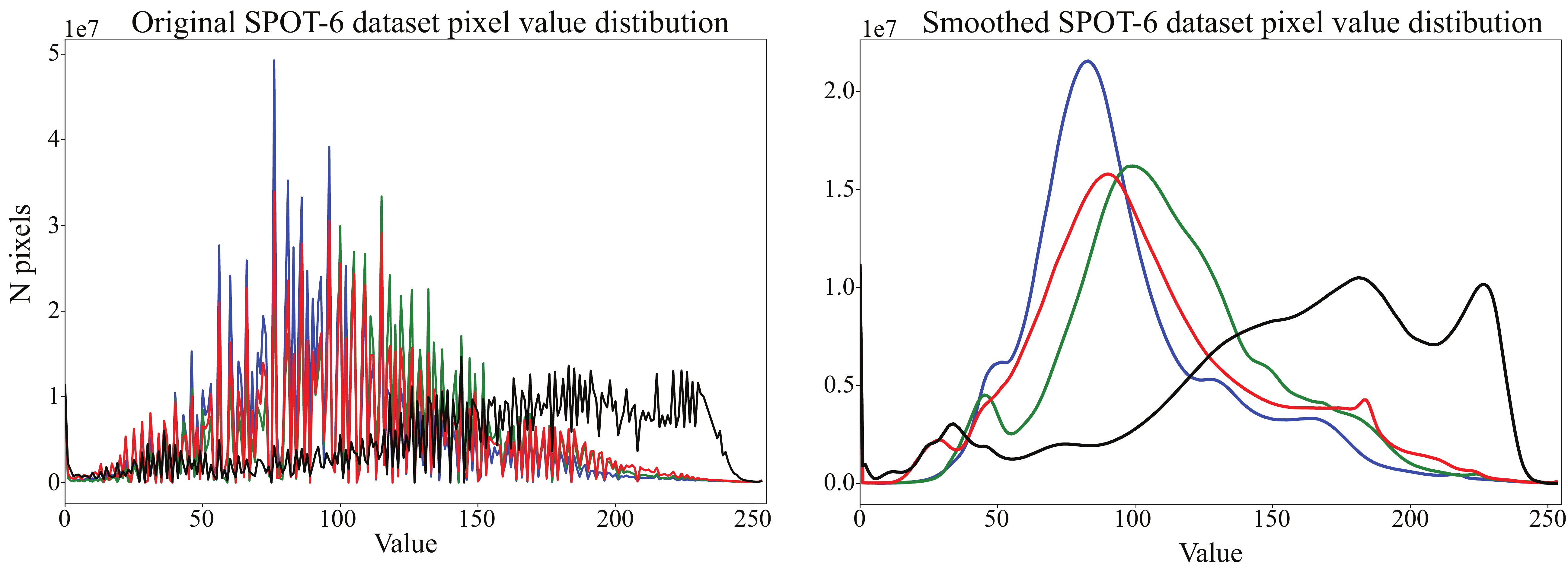}
	\caption{The histogram of pixel value distribution of the original target domain dataset before and after applying a smoothing filter. The blue, red, green, and black lines represent a number of pixel values in the blue, red, green, and near IR channels of the entire dataset, respectively.}
	\label{fig:spot6distr}
\end{figure}

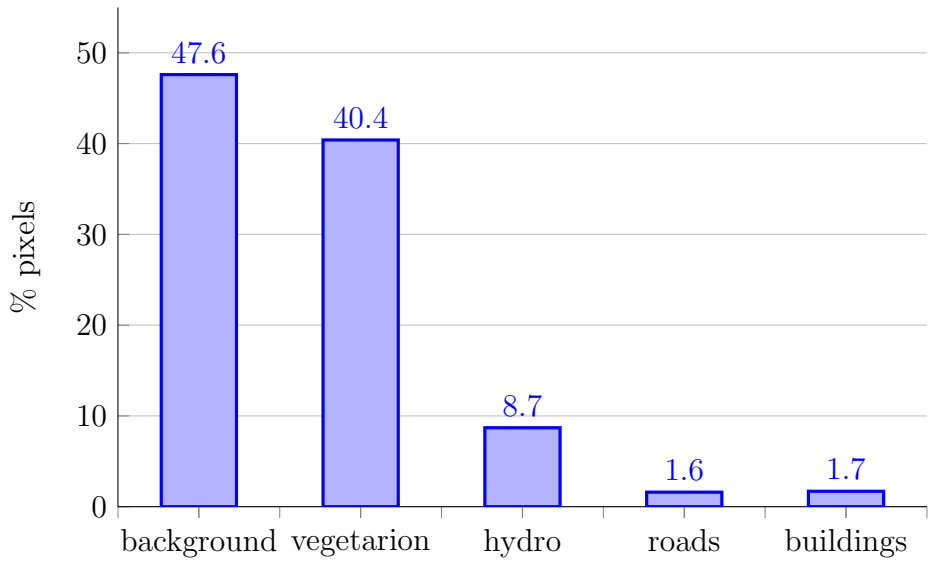
\begin{figure}[!ht] 
\centering
\begin{tikzpicture} 
    \begin{axis}[
        ybar,
        xlabel = \empty,
        xmin = 0.5,
        xmax = 5.5,
        ymin = 0,
        ymax = 55,
        axis x line* = bottom,
        axis y line* = left,
        ylabel= \% pixels,
        width= 0.9\textwidth,
        height = 0.6\textwidth,
        ymajorgrids = true,
        bar width = 10mm,
        xticklabels = \empty,
        nodes near coords,
        extra x ticks = {1,2,3,4,5},
        extra x tick labels = {background, vegetarion, hydro, roads, buildings},
        ]
        \addplot+[mark=none, very thick] coordinates {
            (1,47.6)
            (2,40.4)
            (3,8.7)
            (4,1.6)
            (5,1.7)
        };
    \end{axis} 
\end{tikzpicture}
\caption[Label distribution in the target dataset.]{Label distribution in the target dataset.}
\label{fig:targetlbl}
\end{figure}

SPOT-6 imagery was used as a target domain or the domain where labels do not participate in training the model and which style must be transferred to the source images. This dataset is a preprocessed 1.5-meter 8-bit set of images taken in spring, summer and fall months across Canada. The GeoTIFF raster images and corresponding label geopackages also were cropped into samples of size 4 $\times$ 512 $\times$ 512 and 1 $\times$ 512 $\times$ 512, respectively, the same as for the source dataset. Since the originally provided SPOT-6 imagery had a spiky pixel value distribution (as shown in Fig. \ref{fig:spot6distr}), the whole dataset was smoothed using a Gaussian filter \citep{gauss} with the {\em sigma} parameter equal to (1, 1, 0) and the {\em order} to 0. The original pixel value distribution and the resulting distribution are depicted in Fig. \ref{fig:spot6distr}. The total number of target samples is 4735, with the label distribution given in Fig. \ref{fig:targetlbl}.

\subsection{Training parameters}
\label{section:params}
%\subsection{Noisy labeller}
%\label{section:paramsnoisy}
%Prior to the domain adaptation model training, the noisy labeller was trained on the source domain images with the corresponding labels. The mean of each band for the whole dataset was calculated first, then subtracted from each image during the training. As an optimizer, the Adam method was chosen. The initial learning rate and weight decay were set to $1\times 10^{-4}$ and $5\times 10^{-4}$, respectively. During training, the learning rate was decreased using polynomial decay with a power of 0.9 with the formula:
%\begin{equation}\label{eq:poly_weight_decay}
%lr_{c} = lr_{b}\times \left(1 - \frac{iter_{c}}{iter_{m}}\right)^{p},
%\end{equation}
%where $lr_c$ is the current learning rate, $iter_c$ is a current training iteration, $iter_m$ is a maximum number of training iterations, and $p$ is a decaying power. The model was trained with 8 images in a batch, over 90000 steps. 

\subsection{Domain adaptation model}
\label{section:daparams}
The domain adaptation model was trained with the following parameters. The generator of the model was optimized by the Adam optimizer with parameters $beta1$ and $beta2$ equal to 0.5 and 0.999, respectively. The initial learning rate was equal to $10^{-4}$ and was decayed over training steps with the formula
\begin{equation}\label{eq:daugnet_weight_decay}
lr_{c} = lr_{b}\times \frac{iter_{m} - iter_{c}}{iter_{m} - iter_{d}},
\end{equation}
where $lr_c$ is the current learning rate, $iter_c$ is the current training iteration, $iter_m$ is the maximum number of training iterations, and $iter_d$ is the number of iterations where decaying starts. The total number of training steps was 100,000, and the learning rate decaying step was 75,000. The discriminator of the model was optimized using the Adam optimizer with the same parameters as were used for the generator's optimizer except for the initial learning rate, which was set to $10^{-5}$. Since the style-transferring part of the objective function has many composing losses, the following coefficients were assigned to each of them. The adversarial losses for the generator (in both directions) are multiplied by 1, the cross-reconstruction loss is multiplied by 20, the self-reconstruction loss is multiplied by 10, and the edge loss is multiplied by 25. The images from both domains were normalized from (-1, 1) and then packed in batches of size 1.

\subsection{Segmentation model}
\label{section:segparams}
After the domain adaptation model is trained, all source domain images are transferred to the style of the target domain. A segmentation model was used to evaluate the quality of the translated images. It was represented by the DeepLab v2 framework \citep{chen_deeplab} with a modified number of input channels equal to 4. Same as for the SemI2I method, the original source images were mixed with their stylized versions and used as a training dataset. Training the model using only stylized images did not improve the segmentation performance significantly. As an optimizer, the Adam method was chosen. The initial learning rate and weight decay were set to $1\times 10^{-4}$ and $5\times 10^{-4}$, respectively. During training, the learning rate was decreased using polynomial decay with a power of 0.9 using
\begin{equation}\label{eq:poly_weight_decay}
lr_{c} = lr_{b}\times \left(1 - \frac{iter_{c}}{iter_{m}}\right)^{p},
\end{equation}
where $lr_c$ is the current learning rate, $iter_c$ is a current training iteration, $iter_m$ is a maximum number of training iterations, and $p$ is a decaying power. The model was trained with 8 images in a batch, over 90,000 steps. The validation was performed over the whole target dataset. Prior to the segmentation model training, the mean of each band for all datasets (training and validation) were calculated first, and then subtracted from each image during the training.
\section{Results}
\label{section:results}
Results of the proposed DA model were compared with SemI2I and CyCADA models, and also with the baseline model. The baseline model is a segmentation model trained on data that was not adapted. In this case, the baseline model was trained strictly on source data and then validated on the target dataset. It has the same architecture and the same training parameters as the model discussed in Sec. \ref{section:segparams}.

\begin{figure}[!ht]
	\centering
	\includegraphics[width=\textwidth,keepaspectratio]{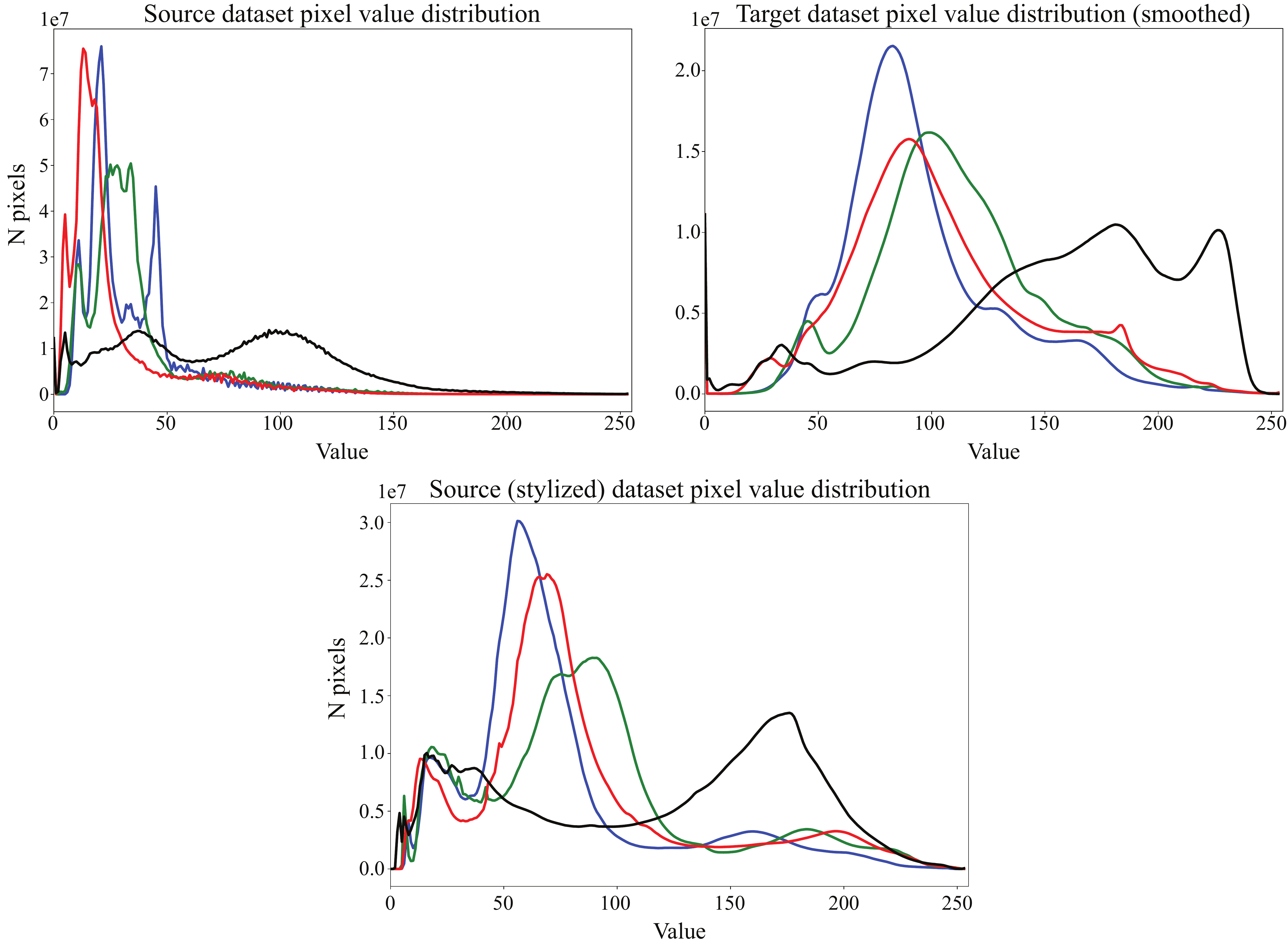}
	\caption{The original source, target and stylized source pixel value distribution. The blue, red, green, and black lines represent a number of pixel values in the blue, red, green, and near IR channels of the entire dataset, respectively.}
	\label{fig:figdistr}
\end{figure}

The numerical results of the style transferring phase are presented in Fig. \ref{fig:figdistr}. The original source dataset pixel values are primarily grouped between 0 and 50, with the near IR band values peak at 100. The original (smoothed) target pixel values are between 50 and 150, with the near IR band values peaking at around 175. The bottom-most plot shows the stylized source dataset where pixel values look similar to the target pixel value distribution.

\begin{figure}[!ht]
	\centering
	\includegraphics[width=\textwidth,keepaspectratio]{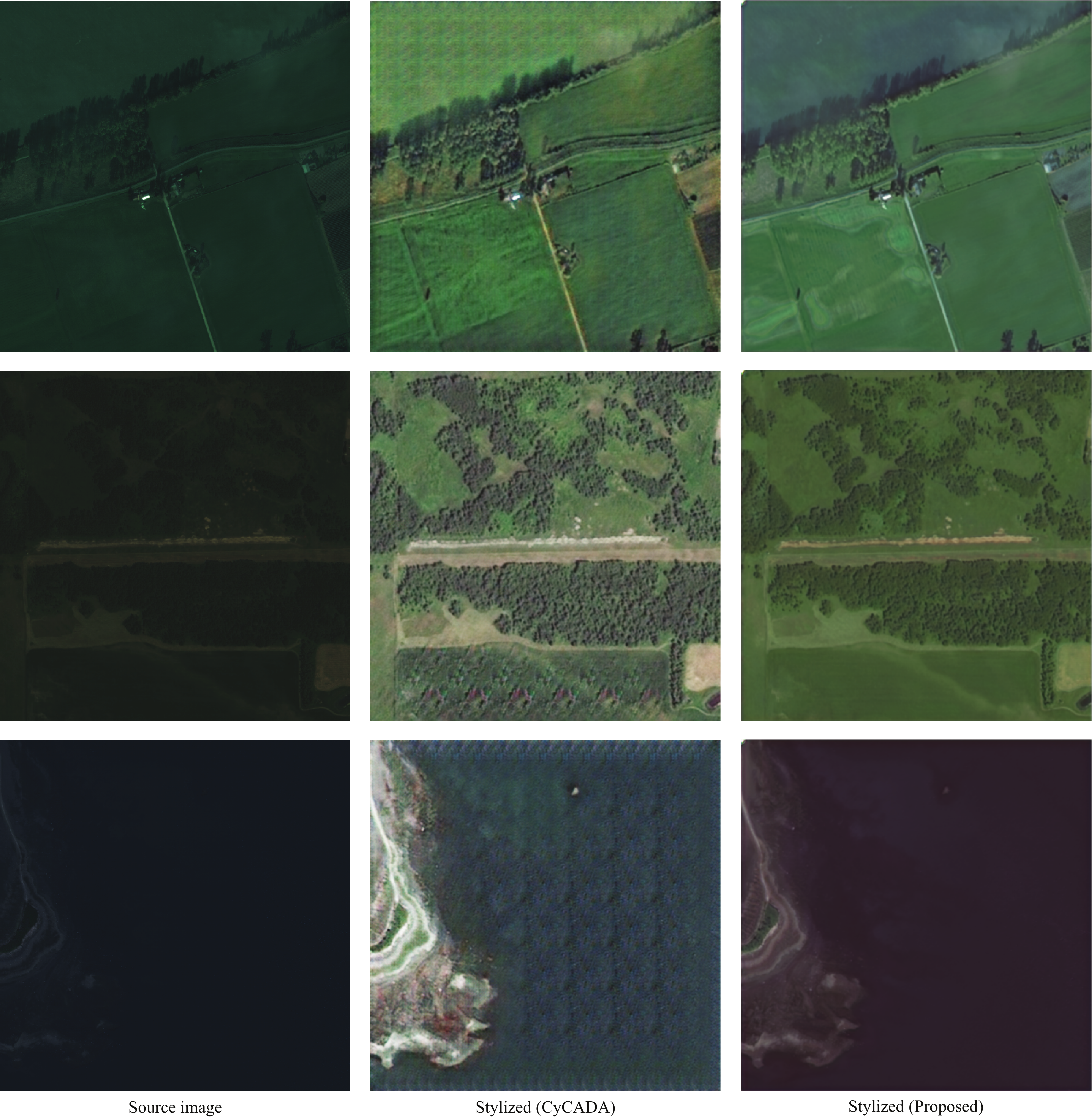}
	\caption{The original source image, its stylized version by CyCADA, and its stylized version by the proposed model.}
	\label{fig:figpatterns}
\end{figure}

As can bee seen from Table \ref{table:experinment_results}, the proposed HRSemI2I method demonstrates performance improvement, both overall ($\sim$3\% to 10\%) and per-class compared to the SemI2I model and the baseline model. Moreover, the performance is comparable with the CyCADA model, which is considered the state-of-the-art domain adaptation method. 
\begin{table}[h!]
\resizebox{\textwidth}{!}{%
\centering
\begin{tabular}{||l| c| c c c c c||}
% \hline
% \multicolumn{3}{|c|}{Country List} \\
 \hline
\textbf{Model} & mIoU & background & vegetation & hydro & roads & buildings\\
 \hline
 \multicolumn{7}{||l||}{\textbf{No Adaptation}} \\
 \hline
 Baseline & 53.02 & 69.48 & 68.77 & 73.78 & 36.63 & 16.43 \\
 \hline
 \multicolumn{7}{||l||}{\textbf{Unsupervised Domain Adaptation}} \\
 \hline
 \hline 
 SemI2I & 60.25 & 74.80 & 76.81 & 77.34 & 42.18 & 30.11 \\
 \hline 
 CyCADA & 63.92 & 75.47 & 75.60 & 81.52 & 46.67 & 40.35 \\
 \hline 
 Proposed & 63.99 & 76.46 & 80.39 & 80.35 & 44.96 & 37.38 \\
 \hline 
\end{tabular}}
\caption[The models' performance (IoU).]{The models' performance (IoU).}
\label{table:experinment_results}
\end{table}
Even though numerically the proposed and CyCADA models produce similar metric values, visually, the proposed model has a significant advantage. As can be seen in Fig. \ref{fig:figpatterns}, some of the style-transferred images generated by the CyCADA model have a noticeable pattern structure. In upper-most example, the water body was replaced by square patches, which look like a green field. This could be explained due to the original source image's green appearance and a noisy labeller marked the water body as a green field, which led to semantically inconsistent style translation. Similar pattern structure can be seen in the middle and bottom-most examples. The outputs of the proposed model, however, demonstrates a free-of-patterns structure with preserved semantic meaning for the objects.

The role of semantic consistency becomes even more evident when applied to satellite images with high spatial resolution. The application of high spatial resolution images to classify fine objects such as cars, roads, individual trees, {\em etc.}, will require perfect style translation where all semantic objects preserve their original meaning. With the growing fleet of high-resolution satellites, this solution becomes even more relevant.

\section{Conclusion}
\label{section:conclusion}
Big earth data generated from a number of satellites is now available, however, the labelling process for these images is expensive and time-consuming. Moreover, multi-generational satellites such as the Landsat constellation may have multiple years of labelled data where more recent satellites may not currently have any labelled data publicly available data. Therefore, unsupervised domain adaptation methods may be used to facilitate the annotating of the images. One of the most recent domain adaptation approaches is style-transferring when the style of the target domain transfers to the source domain images. However, it is essential to provide semantic consistency and per-pixel accuracy during the style-transferring process because each pixel in the RS image is meaningful.

The proposed model presented here improves the previously developed idea of using an adaptive instance normalization layer, maintaining semantic consistency and per-pixel accuracy for the style transferred images. The results were compared to the
state-of-the-art CyCADA model, never applied to RS applications, using the WorldView-2 and SPOT-6 datasets. The results of the proposed model are comparable to the results of the CyCADA model, however, our model is better at preserving semantic consistency which is essential for RS applications.

The future development of the proposed method could include ecological domain transfer, {\em a priori} evaluation of dataset quality in terms of data distribution, or exploration of residual blocks in the encoder. First, both datasets (the source and the target) are composed of satellite images taken in different ecological regions and, thus, represent different style characteristics. Potentially, there could be a situation when the target domain is represented by so many samples taken from different regions that the task of acquiring its {\em common} style becomes meaningless because of the increased complexity of the style-transferring task. A region-to-region domain adaptation approach can lower the complexity and increase the accuracy of the style-transferring task. Second, the target dataset was smoothed by Gaussian filter to overcome its spiky pixel value distribution. The next step could be replacing the target dataset with one where the pixel value distribution is initially smoothed and evaluating the segmentation performance in this case. Third, the proposed DA model distinguishes from the SemI2I by different encoder and decoder structures of the generator. However, the last three residual blocks of the encoder remain unchanged throughout the methods. As was explored by \citet{hong_condgen}, the number of the residual blocks in the generator can substantially influence the domain adaptation results. Therefore, changing the number of the residual blocks is worth further exploration.

%% The Appendices part is started with the command \appendix;
%% appendix sections are then done as normal sections
%\appendix

%% \section{}
%% \label{}

%% For citations use: 
%%       \citet{<label>} ==> Jones et al. [21]
%%       \citep{<label>} ==> [21]
%%

%% If you have bibdatabase file and want bibtex to generate the
%% bibitems, please use
%%
\bibliographystyle{elsarticle-num-names} 
\bibliography{references}
\newpage

%% else use the following coding to input the bibitems directly in the
%% TeX file.
%
%\begin{thebibliography}{00}

%% \bibitem[Author(year)]{label}
%% Text of bibliographic item
%
%\bibitem[ ()]{}
%
%\end{thebibliography}
\end{document}